\documentclass[sigconf]{acmart}

\copyrightyear{2026}
\acmYear{2026}
\setcopyright{cc}
\setcctype{by}
\acmConference[SIGIR '26]{Proceedings of the 49th International ACM SIGIR Conference on Research and Development in Information Retrieval}{July 20--24, 2026}{Melbourne, VIC, Australia}
\acmBooktitle{Proceedings of the 49th International ACM SIGIR Conference on Research and Development in Information Retrieval (SIGIR '26), July 20--24, 2026, Melbourne, VIC, Australia}
\acmISBN{979-8-4007-2599-9/2026/07}
\acmDOI{10.1145/3805712.3809852}
\usepackage{multirow}
\usepackage{balance}

\begin{document}
\title{Diagnosing LLM Reranker Behavior Under Fixed Evidence Pools}

\author{Baris Arat}
\email{baris.arat@ozu.edu.tr}
\affiliation{%
  \institution{Ozyegin University}
  \city{Istanbul}
  \country{Turkey}
}

\author{Emre Sefer}
\email{emre.sefer@ozyegin.edu.tr}
\affiliation{%
  \institution{Ozyegin University}
  \city{Istanbul}
  \country{Turkey}
}
\authornote{Corresponding author.}

\renewcommand{\shortauthors}{Baris Arat and Emre Sefer}



\begin{abstract}

Standard reranking evaluations study how a reranker orders candidates returned by an upstream retriever. This setup couples ranking behavior with retrieval quality, so differences in output cannot be attributed to the ranking policy alone. We introduce a controlled diagnostic for reranking that uses Multi-News clusters as fixed evidence pools. We limit each pool to eight documents and pass identical inputs to all rankers. Within this setup, BM25 and MMR serve as interpretable reference points for lexical matching and diversity optimization. Across 345 clusters, we find that redundancy patterns vary by model: one LLM implicitly diversifies at larger selection budgets, while another increases redundancy. In contrast, LLMs underperform on lexical coverage at small selection budgets. As a result, LLM rankings diverge substantially from both baselines rather than consistently approximating either strategy. By reducing retrieval variance through fixed pools, we interpret these differences more directly as differences in ranking policy. This diagnostic is model agnostic and can be applied to any ranker, including open source systems and proprietary APIs. Our code and processed data for both the Multi-News diagnostic and the complementary TREC-DL evaluation are publicly available at \url{https://github.com/aratbaris/llm_reranker_multinews.git}.

\end{abstract}


\maketitle

\section{Introduction}

LLMs have recently been studied as rerankers in multi-stage retrieval pipelines, and recent toolkits facilitate experimentation with these approaches~\cite{nogueira2020documentrankingpretrainedsequencetosequence, sun_is_2024, ma_zero-shot_2023, pradeep_rankvicuna_2023, sharifymoghaddam_rankllm_2025}. Although LLM methods perform well in benchmarks, benchmark scores provide limited insight into model behavior. Do LLM rerankers behave like lexical matchers, essentially approximating BM25 with better term understanding? Do they implicitly diversify and trade off relevance against redundancy the way MMR does? Or do they follow patterns we cannot yet name? Recent surveys identify interpretability as a key open challenge for LLM-based retrieval methods~\cite{zhu_large_2025}, and answering these questions requires analysis beyond aggregate performance.

In TREC-style evaluation, relevance judgments are pooled from top-ranked results of multiple systems and support comparative evaluation of ranked outputs~\cite{voorhees_2002_phil_ir, buckley_2004_ir_phil}. More broadly, learning to rank benchmarks and neural reranking studies evaluate ranking functions over candidate sets supplied by an initial retrieval stage~\cite{qin_letor_2010, nogueira2020passagererankingbert,zuccon_r2_2025}. This setup reflects real retrieval pipelines, but it also means that reranking behavior is observed only after the retriever has shaped the candidate pool. When two rerankers produce different outputs, the difference may come from ranking policy, from the composition of the candidate set, or from both. Standard reranking evaluation therefore does not by itself support clean attribution of observed differences to reranking policy alone.

To better isolate ranking behavior from candidate generation, we study reranking on fixed candidate pools built from Multi-News clusters~\cite{fabbri_multi-news_2019}. Each cluster contains a small set of related news articles together with an editor-written summary. We use the first sentence of that summary as the query because it often states the main point of the cluster, and we ask how documents within this small topical pool should be ordered with respect to that query statement. In this setting, the query functions more like a summary lead than a traditional retrieval query, and document utility is evaluated within a small editor-selected evidence pool rather than a retriever-produced candidate set. This design lets us analyze ranking behavior through set-level properties such as coverage, redundancy, and semantic support in addition to ranking agreement. We treat this setup as a controlled behavioral diagnostic that complements, rather than replaces, standard IR evaluation, and we also report a complementary TREC-DL reranking experiment to compare patterns across the two settings.

Because our diagnostic only requires the ranker to produce an ordering, it works with black-box models. We do not need access to model weights, attention patterns, or internal representations. This makes the method equally applicable to open source models that we run locally and to proprietary APIs where we can only observe inputs and outputs.

We test two hypotheses about LLM ranking behavior. The first is that LLM rerankers behave like lexical matchers and produce rankings similar to BM25. The second is that they act like implicit diversifiers and produce rankings similar to MMR. We evaluate three LLM rankers alongside BM25, MMR, and random ordering. BM25~\cite{robertson_probabilistic_2009} and MMR~\cite{carbonell_use_1998} serve as landmarks because their objectives are explicit. Such policy-based comparisons have precedent in diversified ranking~\cite{zhai_beyond_2003} and in the comparative evaluation of multi-document summarization methods~\cite{fabbri_multi-news_2019, roy_review_2024}.

Our results reject the first hypothesis and partially support the second. LLM rankings correlate weakly with BM25 (Kendall $\tau$ between 0.19 and 0.41 depending on the model). On lexical coverage, LLMs consistently fall short of both BM25 and MMR. On redundancy, behavior differs across models: Llama implicitly diversifies at larger budgets, while GPT trends toward higher redundancy, and Qwen lies in between. These patterns hold for both lexical and semantic redundancy measures. This variation suggests that LLM ranking behavior cannot be reduced to a single interpretable policy. Our contributions can be summarized as follows:

\begin{itemize}
    \item We propose a controlled diagnostic for reranking analysis that isolates ranking policy from retrieval variance through fixed and topically coherent evidence pools.
    \item We implement this diagnostic with Multi-News clusters, using them as human-curated evidence pools for studying within-pool ranking behavior rather than as a substitute for standard corpus-scale IR evaluation.
    \item We provide empirical evidence that LLM rerankers diverge from lexical baselines and exhibit model-specific redundancy patterns.
\end{itemize}

\section{Methods}

\begin{figure}[t]
\centering
\includegraphics[width=0.9\columnwidth]{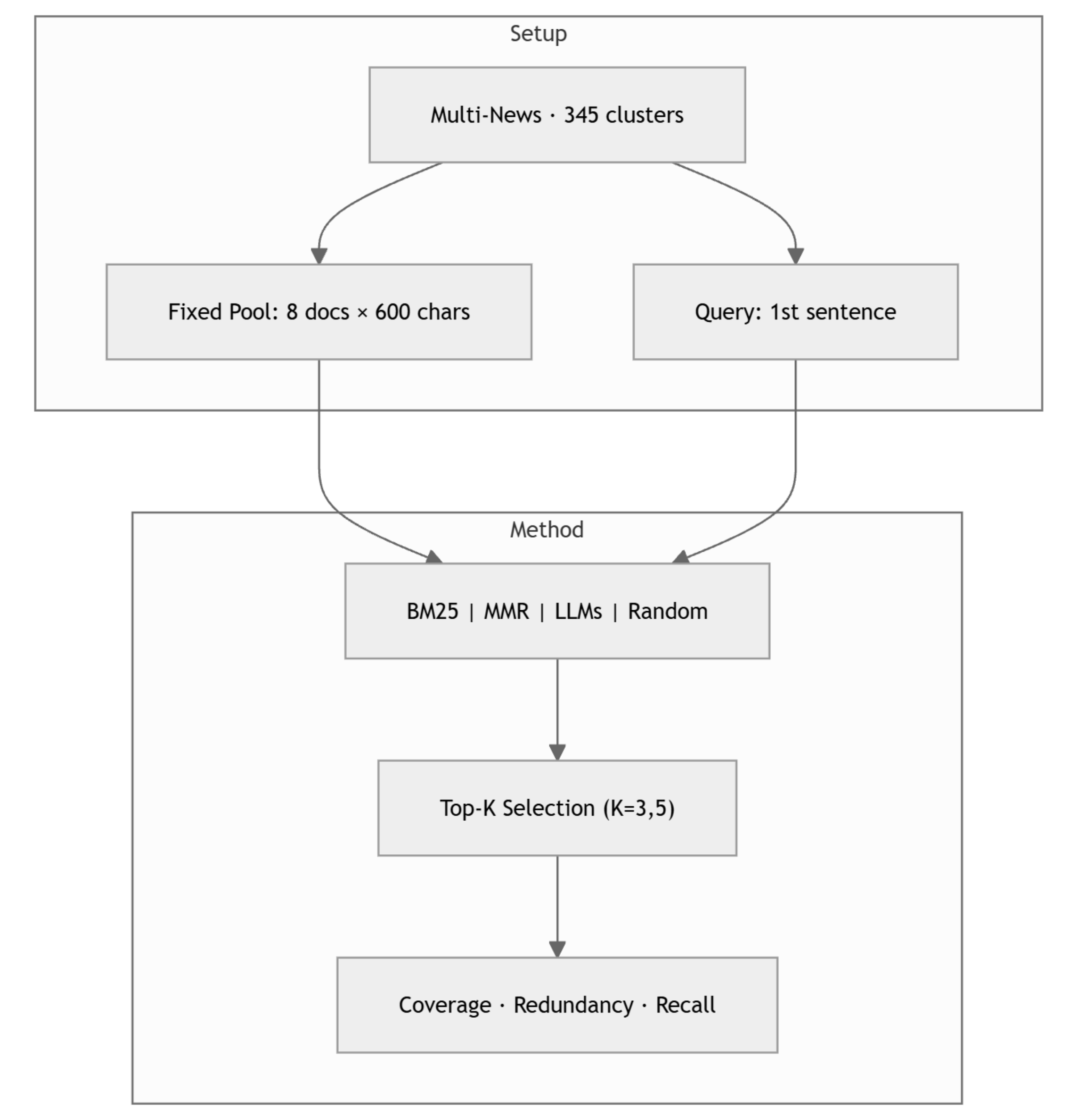}
\caption{Experimental setup with fixed evidence pools.}
\Description{Pipeline diagram showing the experimental setup: Multi-News clusters are used to form fixed pools of eight documents and a query derived from the first summary sentence; multiple rankers produce rankings that are evaluated using coverage, redundancy, and recall metrics.}
\label{fig:pipeline}
\end{figure}

Figure~\ref{fig:pipeline} summarizes the fixed-pool diagnostic setup used throughout the experiments.

\subsection{Problem Definition}

Let $\mathcal{P} = \{d_1, \ldots, d_n\}$ denote a fixed evidence pool of $n$ documents and $q$ a query. A ranking policy $\pi$ maps $(q,\mathcal{P})$ to a strict total ordering of the documents in $\mathcal{P}$, equivalently an ordering of their indices $\{1,\ldots,n\}$. Given a selection budget $K$, $S_K^\pi \subseteq \mathcal{P}$ denotes the unordered set of the $K$ highest-ranked documents under $\pi$. Our diagnostic compares selections $S_K^\pi$ across different ranking functions $\pi \in \{\text{BM25}, \text{MMR}, \text{Random}\} \cup \Pi_{\text{LLM}}$ using coverage and redundancy metrics computed over the same pool $\mathcal{P}$. Here $\Pi_{\text{LLM}} = \{\text{Llama}, \text{Qwen}, \text{GPT}\}$ denotes the set of LLM-based ranking functions evaluated in this study.

\subsection{Data}

We use Multi-News~\cite{fabbri_multi-news_2019}, a dataset of news article clusters created by professional editors for multi-document summarization. Each cluster groups sources that an editor judged relevant to a single topic. Since our study does not involve hold-out experiments, we merge the original train, validation, and test splits while retaining clusters with at least eight usable sources. Multi-News clusters contain up to ten source documents, but clusters with nine or ten usable sources are less common. Choosing eight documents preserves a relatively large pool size while retaining 345 clusters in total. For clusters containing more than eight sources, we deterministically select eight using a fixed seed. We use the first sentence of the editorial summary as the query and truncate it to 400 characters. Each document is also truncated to 600 characters before it is passed to a ranker. These character limits are used only to standardize the text input and token processing is defined separately for lexical baselines and metrics. Table~\ref{tab:params} summarizes these choices.

\begin{table}[t]
\centering
\small
\caption{Pool standardization parameters.}
\label{tab:params}
\begin{tabular}{ll}
\hline
Parameter & Value \\
\hline
Documents per pool & 8 documents \\
Query source & First sentence of summary \\
Query length limit & 400 characters \\
Document length limit & 600 characters \\
\hline
\end{tabular}
\end{table}

\subsection{Rankers}
We compare six ranking approaches. Queries and documents are truncated by character count as described above. For lexical baselines and lexical metrics, content tokens are extracted separately by lowercase regex matching of alphanumeric sequences, with stopword removal and digit filtering. \textbf{BM25} ranks documents by lexical matching to the query~\cite{robertson_probabilistic_2009}, implemented via \texttt{rank-bm25}~\cite{rank_bm25} with default parameters. \textbf{MMR}~\cite{carbonell_use_1998} starts from relevance scores (BM25 in our case) and performs greedy selection to reduce redundancy, using Jaccard similarity between document content-token sets with $\lambda = 0.7$. As a baseline, \textbf{Random} ranks documents in random order using a fixed seed.

For \textbf{LLM rankers}, we test three instruction-tuned models: Llama 3.1 8B Instruct~\cite{grattafiori_llama3_2024}, Qwen2.5 7B Instruct~\cite{qwen25_2024}, and GPT-5-mini (accessed via the OpenAI Chat Completions API, model=gpt-5-mini, other parameters left at API default)~\cite{singh2025openaigpt5card}. Each model receives the query and all eight documents in a single prompt, then returns a complete ranking in JSON format. We shuffle document presentation order deterministically per cluster using a fixed seed, and the same order is shared across all LLM models. This controls for position bias, as LLMs are known to favor documents at certain positions independently of document relevance~\cite{tang-etal-2024-found, liu-etal-2024-lost}. The prompt instructs the model to rank by relevance only:

\begin{quote}
\small\ttfamily
You are ranking evidence documents for a query.\\
Task: You are given exactly 8 candidate documents.\\
Rank ALL 8 documents from best to worst.\\
Ranking goal: 1) Relevance to the query.\\
Return ONLY strict JSON with exactly 8 unique indices:\{"ranked\_indices":[...]\}
\end{quote}

In our experiments, the parsing fallback rate was 0.0\% across all runs. For open source models, we set the temperature to zero. For the API model we use the default configuration, since this parameter cannot be overridden in the current version of the API. We compute metrics for multiple selection budgets $K \in \{3,4,5,6\}$. The results follow the same trends across these settings. Table~\ref{tab:bootstrap_all} reports $K=3$ and $K=5$ as representative small and medium budgets, while Figure~\ref{fig:metrics} shows all four values to illustrate trends across selection sizes.

\subsection{Metrics}
In the Multi-News setting, we evaluate not only ranking agreement but also how well a selected set covers the query and reference summary while avoiding redundant evidence. These metrics are diagnostic rather than direct relevance judgments, and we use them to compare how ranking policies behave within the fixed pool.

For a ranker $\pi$ and selection budget $K$, let $S_K^\pi$ denote the unordered set of top-$K$ selected documents. Let $Q$ denote the set of content tokens extracted from the query, $R$ the set of content tokens in the reference summary, and let
\[
T_K^\pi = \bigcup_{d \in S_K^\pi} \mathrm{Tok}(d)
\]
denote the union of content tokens across the selected documents, where $\mathrm{Tok}(d)$ extracts content tokens from document $d$ and applies the same procedure to documents, queries, and summaries.

\textbf{Lexical coverage} measures what fraction of query content appears in the selected documents:
\[
\text{coverage} = \frac{|T_K^\pi \cap Q|}{|Q|}
\]

\textbf{Lexical redundancy} captures overlap among selected documents. For each pair of selected documents $d_i, d_j \in S_K^\pi$, let $D_i=\mathrm{Tok}(d_i)$ and $D_j=\mathrm{Tok}(d_j)$. We compute Jaccard similarity and average over all pairs:
\[
\text{redundancy} = \frac{1}{\binom{K}{2}} \sum_{i < j} \frac{|D_i \cap D_j|}{|D_i \cup D_j|}
\]
where the sum ranges over all unordered pairs in $S_K^\pi$.

\textbf{Summary recall} measures how much of the reference summary the selection covers lexically:
\[
\text{summary-recall} = \frac{|T_K^\pi \cap R|}{|R|}
\]

For semantic metrics, we encode documents and summary sentences using the MiniLM-L6-v2 model~\cite{reimers-2019-sentence-bert}. Each selected document $d_i \in S_K^\pi$ is represented by an embedding vector $\mathbf{e}_i$, and each sentence in the reference summary is represented by an embedding vector $\mathbf{s}_m$. All embeddings are L2-normalized, so cosine similarity reduces to the dot product. 

\textbf{Semantic redundancy} mirrors the lexical version over embeddings of selected documents:
\[
\text{sem-redundancy} = \frac{1}{\binom{K}{2}} \sum_{i<j} \mathbf{e}_i \cdot \mathbf{e}_j
\]
where the sum ranges over all unordered pairs in $S_K^\pi$.

\textbf{Semantic coverage} measures how well the selection supports each sentence in the reference summary at the sentence level. We split summaries into sentences using regex on terminal punctuation. For each summary sentence embedding $\mathbf{s}_m$, we take the maximum similarity to any selected document, then average across sentences:
\[
\text{sem-coverage} = \frac{1}{M} \sum_{m=1}^{M} \max_{d_j \in S_K^\pi} (\mathbf{s}_m \cdot \mathbf{e}_j)
\]
where $M$ is the number of summary sentences.

We compute Kendall $\tau$ between the full rankings over the same 8-document pool. Top-3 Jaccard is $|S_3^{\pi_1}\cap S_3^{\pi_2}|/|S_3^{\pi_1}\cup S_3^{\pi_2}|$. We report paired bootstrap differences with 95\% confidence intervals over 10,000 resamples.

\subsection{Complementary TREC-DL Evaluation}

To connect the Multi-News diagnostic to a standard IR setting, we run a complementary reranking experiment on pooled TREC-DL 2021--2023 queries. We retain queries with at least five highly relevant passages, which yields 123 comparable queries with 100 passages each. This choice lets us study small selection budgets over pools that contain sufficient strongly relevant passages. In this setting, we evaluate BM25, MMR, a pretrained CrossEncoder reranker, \texttt{ms-marco-MiniLM-L6-v2}~\cite{reimers-2019-sentence-bert, ce_msmarco_minilm_l6_v2}, GPT-5.4-mini~\cite{openai_gpt54mini_2026}, Qwen, Llama, and Random on top-3 and top-5 selection. By design, this experiment is not a direct replication of the Multi-News setup. It serves instead as an independent evaluation in a pooled IR setting with passage relevance labels. We report effectiveness metrics such as precision, recall, hit rate, and nDCG together with pairwise top-$K$ agreement statistics. The GPT model in this experiment comes from the same API and model family as in the Multi-News study, but it uses the version available at evaluation time. Full implementation details, prompts, and preprocessing steps are provided in the accompanying code release.

\section{Results}

Table~\ref{tab:agreement} shows how much each LLM agrees with the baselines. The agreement varies considerably across models. \texttt{Llama} correlates only weakly with BM25 ($\tau = 0.19$), while \texttt{GPT} shows moderate correlation ($\tau = 0.41$). All three LLMs differ clearly from random ordering. These results indicate that LLM rerankers do not simply reproduce lexical rankings, ruling out the first hypothesis.

\begin{table}[t]
\centering
\small
\caption{Ranking agreement averaged over 345 clusters.}
\label{tab:agreement}
\begin{tabular}{lccccc}
\hline
 & \multicolumn{2}{c}{Kendall $\tau$} & \multicolumn{3}{c}{Top-3 Jaccard} \\
\cline{2-3} \cline{4-6}
Model & BM25 & MMR & BM25 & MMR & Rand \\
\hline
\texttt{Llama} & 0.19 & 0.19 & 0.37 & 0.37 & 0.26 \\
\texttt{Qwen}  & 0.33 & 0.32 & 0.42 & 0.41 & 0.25 \\
\texttt{GPT}   & 0.41 & 0.39 & 0.48 & 0.47 & 0.27 \\
\hline
\end{tabular}
\end{table}

\begin{figure}[t]
\centering
\includegraphics[width=0.85\columnwidth]{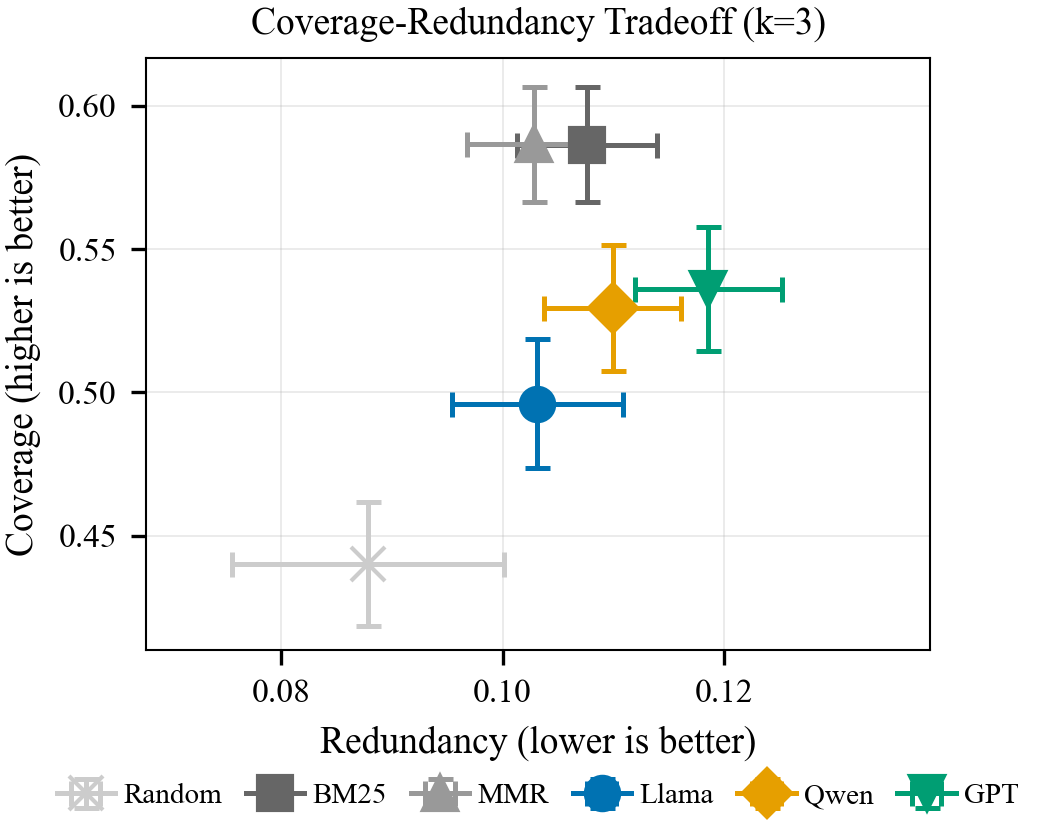}
\caption{Mean coverage and redundancy at $K=3$ with 95\% CIs.}
\Description{Scatter plot showing mean lexical coverage and lexical redundancy at K equals 3 for BM25, MMR, Random, and the LLM rankers, with 95 percent confidence intervals. BM25 and MMR appear in the higher-coverage region, while LLM rankers appear below them.}
\label{fig:tradeoff}
\end{figure}

\begin{figure}[t]
\centering
\includegraphics[width=\columnwidth]{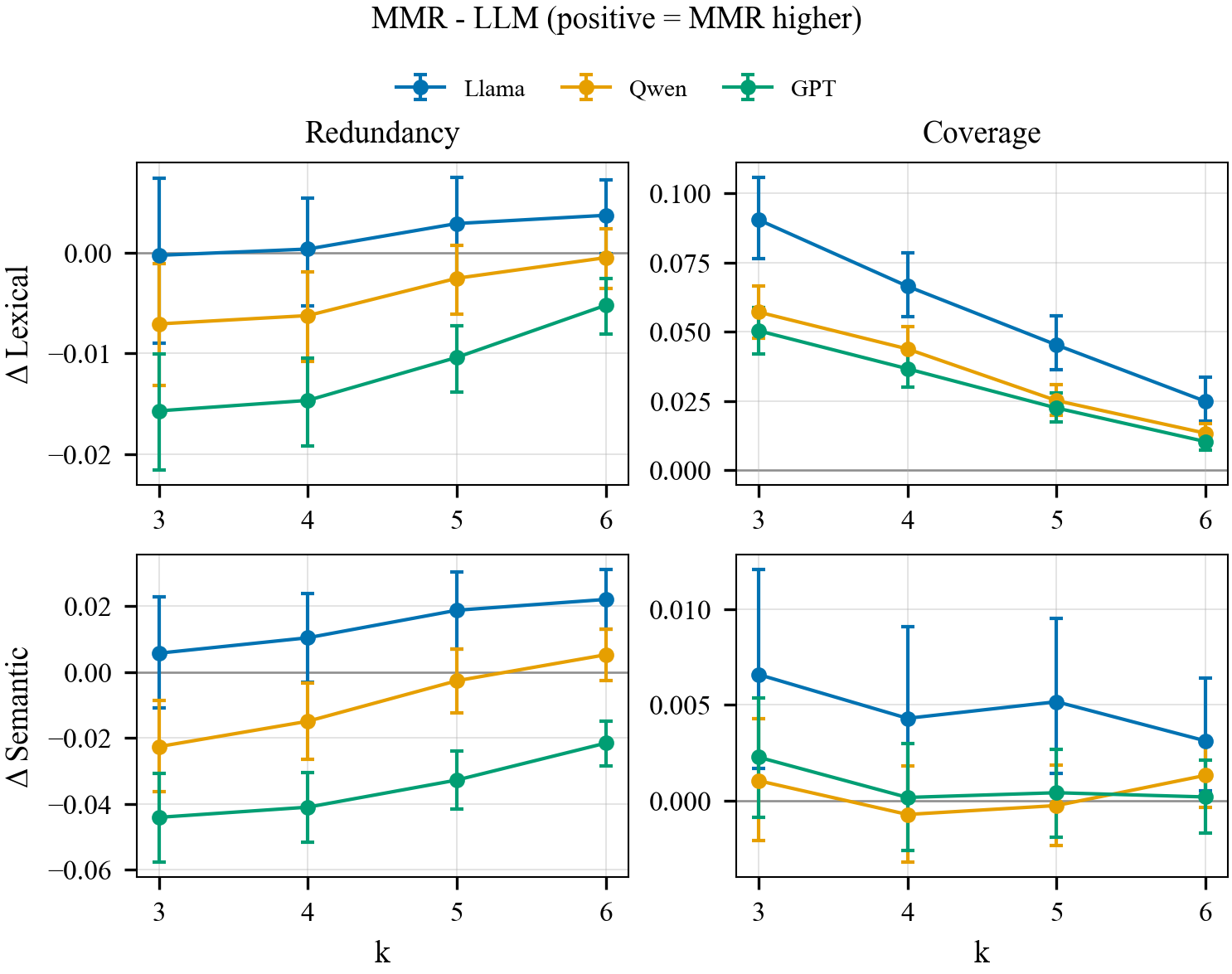}
\caption{Redundancy and coverage deltas (MMR minus LLM) for lexical (top) and semantic (bottom) metrics.}
\Description{Four-panel figure showing MMR minus LLM deltas for lexical and semantic redundancy (left column) and coverage (right column) across $K$ values 3--6. Llama shows increasing redundancy delta trending positive, GPT shows consistently negative redundancy delta, and all models show positive but decreasing coverage delta.}
\label{fig:metrics}
\end{figure}

Table~\ref{tab:bootstrap_all} breaks down the differences by metric and selection budget, and Figure~\ref{fig:metrics} shows all four $K$ values to illustrate trends across selection sizes. Figure~\ref{fig:tradeoff} shows the coverage-redundancy tradeoff at $K=3$, where BM25 and MMR occupy the high-coverage region while LLMs cluster below. We discuss the lexical metrics first, and then the semantic metrics.

\begin{table*}[ht]
\centering
\caption{Paired bootstrap differences (mean $\Delta$, 95\% CI) over 345 clusters. Positive $\Delta$ means the first method is higher; for redundancy, lower is better.}
\label{tab:bootstrap_all}
\small
\setlength{\tabcolsep}{3.5pt}
\begin{tabular}{llcccccc}
\hline
& & \multicolumn{3}{c}{$K=3$} & \multicolumn{3}{c}{$K=5$} \\
\cline{3-5} \cline{6-8}
Model & Metric &
\multicolumn{1}{c}{MMR--LLM} &
\multicolumn{1}{c}{BM25--LLM} &
\multicolumn{1}{c}{LLM--Rand} &
\multicolumn{1}{c}{MMR--LLM} &
\multicolumn{1}{c}{BM25--LLM} &
\multicolumn{1}{c}{LLM--Rand} \\
\hline
\multirow{5}{*}{\texttt{Llama}}
& coverage & +.091 [.077,.106] & +.090 [.076,.106] & +.056 [.037,.076] & +.045 [.036,.056] & +.045 [.036,.056] & +.033 [.019,.047] \\
& redundancy & $.000$ [$-.009$,.007] & +.004 [$-.003$,.012] & +.015 [.004,.025] & +.003 [$-.002$,.008] & +.009 [.004,.014] & +.010 [.004,.015] \\
& summary-recall & +.008 [.002,.014] & +.009 [.003,.015] & +.017 [.010,.025] & +.008 [.003,.014] & +.007 [.002,.013] & +.014 [.008,.020] \\
& sem-redundancy & +.006 [$-.011$,.023] & +.016 [.000,.032] & +.112 [.090,.134] & +.019 [.007,.030] & +.029 [.018,.040] & +.070 [.056,.083] \\
& sem-coverage & +.007 [.002,.012] & +.007 [.002,.012] & +.014 [.008,.020] & +.005 [.001,.010] & +.005 [.001,.009] & +.004 [$-.001$,.008] \\
\hline
\multirow{5}{*}{\texttt{Qwen}}
& coverage & +.057 [.048,.067] & +.057 [.048,.067] & +.089 [.071,.108] & +.025 [.020,.031] & +.025 [.020,.031] & +.053 [.041,.066] \\
& redundancy & $-.007$ [$-.013$,$-.001$] & $-.002$ [$-.008$,.004] & +.022 [.009,.034] & $-.003$ [$-.006$,.001] & +.004 [.000,.007] & +.015 [.010,.020] \\
& summary-recall & +.003 [$-.002$,.008] & +.004 [$-.001$,.009] & +.022 [.015,.030] & +.003 [$-.001$,.008] & +.002 [$-.003$,.007] & +.019 [.013,.024] \\
& sem-redundancy & $-.023$ [$-.036$,$-.009$] & $-.012$ [$-.025$,.001] & +.140 [.118,.162] & $-.003$ [$-.012$,.007] & +.008 [$-.002$,.017] & +.091 [.079,.102] \\
& sem-coverage & +.001 [$-.002$,.004] & +.001 [$-.002$,.004] & +.019 [.013,.026] & $-.000$ [$-.002$,.002] & $-.001$ [$-.003$,.001] & +.009 [.006,.012] \\
\hline
\multirow{5}{*}{\texttt{GPT}}
& coverage & +.050 [.042,.059] & +.050 [.042,.059] & +.096 [.079,.114] & +.023 [.018,.028] & +.023 [.018,.028] & +.056 [.044,.068] \\
& redundancy & $-.016$ [$-.022$,$-.010$] & $-.011$ [$-.017$,$-.005$] & +.031 [.018,.042] & $-.010$ [$-.014$,$-.007$] & $-.004$ [$-.007$,$-.001$] & +.023 [.017,.029] \\
& summary-recall & +.001 [$-.004$,.006] & +.002 [$-.003$,.006] & +.024 [.018,.031] & +.000 [$-.004$,.005] & $-.001$ [$-.005$,.004] & +.021 [.016,.027] \\
& sem-redundancy & $-.044$ [$-.058$,$-.031$] & $-.034$ [$-.046$,$-.021$] & +.162 [.141,.182] & $-.033$ [$-.042$,$-.024$] & $-.023$ [$-.030$,$-.015$] & +.121 [.109,.133] \\
& sem-coverage & +.002 [$-.001$,.005] & +.002 [$-.001$,.006] & +.018 [.012,.025] & +.000 [$-.002$,.003] & $-.000$ [$-.002$,.002] & +.009 [.006,.012] \\
\hline
\end{tabular}
\end{table*}

\textbf{Lexical coverage.} BM25 and MMR consistently achieve higher lexical coverage than the LLM rankers. At $K=3$, the coverage gap ranges from about +0.05 to +0.09 depending on the model, with confidence intervals excluding zero. The gap narrows at $K=5$ but remains significant, at about +0.02 to +0.05. All LLM rankers outperform random selection on coverage, with gains of about +0.06 to +0.10 at $K=3$.

\textbf{Lexical redundancy.} Redundancy patterns differ by model. For \texttt{Llama}, differences at $K=3$ are negligible (MMR--LLM: $-0.000$). At $K=5$, BM25 exhibits higher redundancy than \texttt{Llama} ($+0.009$), with MMR showing a smaller and statistically weaker difference ($+0.003$). This indicates that \texttt{Llama} is consistently less redundant than BM25 and comparable to MMR at larger budgets. For \texttt{GPT}, the pattern reverses: \texttt{GPT} produces more redundant selections than both baselines at both budgets (MMR--LLM: $-0.016$ at $K=3$, $-0.010$ at $K=5$, with confidence intervals excluding zero). \texttt{Qwen} falls in between: slightly more redundant than the baselines at $K=3$ (MMR--LLM: $-0.007$) but comparable at $K=5$. Figure~\ref{fig:metrics} (top left) shows these diverging trajectories with changing selection budget: \texttt{Llama}'s redundancy delta trends upward toward zero and beyond, while \texttt{GPT} remains consistently below the baseline.

\textbf{Summary recall.} Differences in summary recall are smaller but follow the coverage pattern. BM25 and MMR slightly outperform LLMs at $K=3$ (up to $+0.009$ for \texttt{Llama}), though for \texttt{GPT} the difference is small and confidence intervals include zero. All LLMs outperform random ordering ($+0.017$ to $+0.024$).

\textbf{Semantic redundancy.} The bottom left panel of Figure~\ref{fig:metrics} confirms that semantic redundancy follows the same model-specific pattern as lexical redundancy. At $K=3$, \texttt{Llama} shows semantic redundancy comparable to the baselines (BM25--LLM $= +0.016$), while \texttt{GPT} produces substantially more semantically redundant selections (BM25--LLM $= -0.034$, MMR--LLM $= -0.044$). At $K=5$, \texttt{Llama} selections are less semantically redundant than BM25 (BM25--LLM $= +0.029$), consistent with implicit diversification. \texttt{Qwen} again falls between, closer to \texttt{GPT} at $K=3$ but converging with the baselines at $K=5$.

\textbf{Semantic coverage.} Semantic coverage differences are small across all comparisons. BM25 and MMR slightly outperform the LLM rankers ($+0.001$ to $+0.007$), but many confidence intervals include zero, indicating weak effects. The clearest signal is that all LLM rankers outperform random ordering ($+0.014$ to $+0.019$ at $K=3$), which confirms that their selections capture summary content beyond chance.

\textbf{Complementary TREC-DL evaluation.}

We next test whether a similar divergence pattern appears in a TREC-DL pooled reranking setting. On 123 comparable TREC-DL 2021--2023 queries, CrossEncoder and GPT both achieve strong top-$K$ effectiveness, with GPT higher overall. $P@3$ with relevance grade $\geq 2$ as the positive threshold is 0.691 for CrossEncoder and 0.743 for GPT, while nDCG@5 is 0.636 and 0.679, respectively. However, their selected passages overlap only modestly, with mean top-3 and top-5 Jaccard scores of 0.231 and 0.255. BM25 and MMR remain highly similar to each other, with top-3 and top-5 Jaccard scores of 0.820 and 0.821. This setting differs from the Multi-News diagnostic by design and highlights an important point: strong effectiveness can coexist with substantial differences in top-ranked results, and stronger models do not reproduce lexical baselines. Although this evaluation still entangles first-stage retrieval and reranking policy, that limitation is precisely what motivates our Multi-News diagnostic. Together, the two settings suggest that models with similar aggregate effectiveness can reach those results through different ranking behaviors, indicating policy variation.

\section{Conclusion}

We introduced a diagnostic for LLM reranking using human-curated, fixed-size evidence pools. Our analysis complements retrieval benchmarks and should be interpreted as a controlled diagnostic of reranking behavior, rather than as a measure of retrieval performance. It also supports set-level evaluation through coverage, redundancy, and semantic support. Our findings can be summarized in three points. First, LLM rankers do not approximate lexical baselines, with agreement with BM25 ranging from $\tau = 0.19$ to $0.41$ depending on the model. Second, all three LLM rankers incur a coverage cost relative to BM25 and MMR, on the order of about 5--9 percentage points at $K=3$ and about 2--5 at $K=5$. Third, redundancy tradeoffs are model-specific, with \texttt{Llama} tending to diversify at larger selection budgets while \texttt{GPT} consistently produces more redundant selections, and \texttt{Qwen} exhibiting intermediate behavior. Our complementary TREC-DL experiment supports the same qualitative point in a standard pooled IR setting: models with similar effectiveness can still produce substantially different top-ranked selections. These patterns are consistent with broader findings that LLM behavior varies substantially across models~\cite{liang2023holisticevaluationlanguagemodels}.

Stepping back, we view reranking not only as a component of a standard retrieval pipeline, but also as a ranking policy that can be studied under different evaluation settings. In retrieval benchmarks, reranking is evaluated over retriever-produced pools, which is essential for measuring end-to-end effectiveness. In controlled fixed pools such as ours, the same ranking function can be examined with reduced influence from candidate generation, making attribution to ranking policy more direct. Our query and evidence formulation also enables a different evaluation lens based on set properties such as coverage, redundancy, and semantic support. Together, these settings yield complementary insights into model behavior.

Our study leaves several questions open. We test three models under one prompt template, use automatic metrics rather than human judgments, and do not study robustness to prompt variation or document order beyond the shared randomized presentation. We also leave confidence estimation and its relation to ranking quality for future work~\cite{kadavath2022languagemodelsmostlyknow}.

\bibliographystyle{ACM-Reference-Format}
\balance
\bibliography{SIGIR}

\end{document}